\documentclass[10pt, a4paper]{article}

\usepackage{lrec-coling2024} 
\usepackage{amsmath,graphicx}
\usepackage{pifont}
\usepackage{multirow}
\usepackage{booktabs}
\usepackage{enumitem}
\usepackage{subfigure}
\usepackage{makecell}
\usepackage{color}
\usepackage{xcolor}
\usepackage[ruled, lined, longend]{algorithm2e}

\newcommand{\arrowmark}{\ding{233}}
\title{Scaling Data Diversity for Fine-Tuning Language Models \\in Human Alignment}

\name{Feifan Song$^{1}$, Bowen Yu$^{2}$, Hao Lang$^{2}$, Haiyang Yu$^{2}$\\
{\bf \large Fei Huang$^{2}$, Houfeng Wang$^{1}$*\thanks{* Corresponding authors}, Yongbin Li$^{2}$*}} 

\address{$^{1}$National Key Laboratory of Multimedia Information Processing\\School of Computer Science, Peking University\\ 
         songff@stu.pku.edu.cn; wanghf@pku.edu.cn \\ \\
         $^{2}$Alibaba Group \\ 
         {yubowen.ybw, hao.lang, yifei.yhy, f.huang, shuide.lyb}@alibaba-inc.com }

\abstract{
Alignment with human preference prevents large language models~(LLMs) from generating misleading or toxic content while requiring high-cost human feedback. 
Assuming resources of human annotation are limited, there are two different ways of allocating considered: more diverse \textbf{PROMPTS} or more diverse \textbf{RESPONSES} to be labeled. 
Nonetheless, a straightforward comparison between their impact is absent. 
In this work, we first control the diversity of both sides according to the number of samples for fine-tuning, which can directly reflect their influence. 
We find that instead of numerous prompts, more responses but fewer prompts better trigger LLMs for human alignment. 
Additionally, the concept of diversity for prompts can be more complex than responses that are typically quantified by single digits. 
Consequently, a new formulation of prompt diversity is proposed, further implying a linear correlation with the final performance of LLMs after fine-tuning. We also leverage it on data augmentation and conduct experiments to show its effect on different algorithms. 
 \\ \newline  }

\begin{document}

\maketitleabstract
\section{Introduction}
Large Language Models~(LLMs) have gained widespread recognition for their proficiency in many domains, including instruction following, intimation and knowledge utilization~\cite{brown2020language, chung2022scaling, muennighoff2022crosslingual, NEURIPS2022_9d560961, wang2022self, zhou2022large, von2023transformers, dai2023can, yang2023iterative, zhong2023improving, schick2023toolformer, li2023api, song2023restgpt, qin2023toolllm, wang2023making, yang2023not, lyu2024knowtuning}. 
However, they can reveal toxic or offensive content either inadvertently or intentionally, underscoring the importance of aligning them with human values~\cite{bai2022constitutional}. 
The transition of paradigm from model-centric to data-centric~\cite{zha2023data-centric-survey, zha2023data-centric-perspectives} has led to the development of products that are refined using abundant data with human feedback~(e.g., ChatGPT, Claude). 
These products show remarkable capabilities in delivering reliable responses, which prioritizes data collection for LLM fine-tuning aimed at human alignment.

\begin{figure*}[t]
\centering
\includegraphics[width=\textwidth]{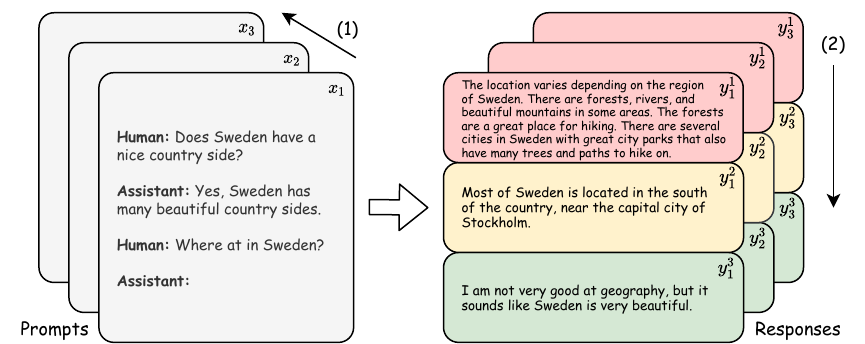}
\caption{Different directions of data expansion for human alignment: (1)~Expanding more prompts; (2)~Expanding more responses for each prompt.}
\label{fig:intro}
\end{figure*}

In this field, a natural challenge is the huge expense of high-quality human annotation for diverse samples~\cite{casper2023open}. 
The greater the diversity within the dataset is, the higher upper bound of performance can be achieved. 
Nevertheless, this diversity also results in higher costs.
In detail, LLMs are forced to generate responses in line with human preference, based on provided prompts. 
When the annotation resources are limited, a decision must be made regarding the allocation of these resources between a broader range of prompts or a larger number of responses to be annotated, as illustrated in Figure~\ref{fig:intro}.

The well-known LLaMA-2~\cite{touvron2023llama2} chooses to utilize samples of human alignment, each containing one prompt and two responses to maximize the prompt diversity.
On the contrary, various studies~\citep{ouyang2022training, yuan2023rrhf, song2023pro} concentrate on providing each prompt with more responses, enabling LLMs to distinguish subtle differences among various candidates.
Although both sides are intuitively reasonable, there is currently a lack of direct comparison and comprehensive analyses between them.

In this paper, we investigate the impact of both prompts and responses for LLM fine-tuning in human alignment. 
We first design a quantitative experiment to assess the effect of the two strategies. 
A series of sub-datasets are created from the raw dataset, some of which contain more prompts but fewer responses for each prompt, while others have more responses for each prompt but fewer prompts.
These subsets maintain a certain proportional relation to maintain a constrained total number of annotations, on which we fine-tune LLMs and compare their performances for comprehensive analyses.

While \citet{song2023pro} has demonstrated the effect of increased responses, a scaling law between prompts diversity and the final performance in human alignment is yet to be established. 
Similar to \citet{kaplan2020scaling} and \citet{muennighoff2023scaling} exploring the correlation token statistics and evaluation metrics, the aforementioned quantitative experiment manipulates prompt diversity by adjusting training set sizes only, overlooking the influence of token combinations representing syntax and contextual information. 
To address this gap, we introduce a novel formulation to empirically define prompt diversity based on N-grams. 
Furthermore, we uncover a linear relationship between this diversity and the acquired reward scores by examining various scales of training sets, different base models, and algorithms.

We also try to enhance data diversity by employing this new formulation to guide a data augmentation process. 
Beginning with existing samples, we sample multiple new prompts and corresponding responses, then assess them based on N-gram overlap with the given demonstrations to determine their acceptance. 
Implementing this method leads to an improvement in performance compared to randomly sampled data.

We conclude from all experiments that:

(1)~Expanding responses yields more benefit than prompts. 
We attribute it to two reasons: just a few prompts can activate LLMs in human alignment, as explained in~\citet{zhou2023lima}, while more responses offer clearer signals for fine-tuning, thus proving more help.

(2)~The empirical formulation of prompt diversity can establish a linear correlation with the final performance of LLMs.

(3)~Directed by the proposed formulation of prompt diversity, the new process of data augmentation can promote the performance of LLMs.

\section{Related Work}
\subsection{Fine-tuning for Human Alignment}
Despite their promising potential, large language models carry the risk of generating toxic or offensive content without human alignment. 
One approach that has gained considerable attention in addressing this issue is Reinforcement Learning from Human Feedback~(RLHF)~\citep{stiennon2020learning, ouyang2022training, bai2022training, zhu2023principled, zhu2023fine, yu2023constructive}. 
For instance, InstructGPT~\citep{ouyang2022training} builds a three-step pipeline of RLHF, which includes supervised fine-tuning~(SFT), reward model~(RM) training, and reinforcement learning using PPO~\citep{schulman2017proximal}.
This process involves collecting numerous samples, each consisting of one prompt and multiple candidate responses ranked by human annotators. 
These annotated rankings are then segmented into pairs to enhance computational efficiency. 
\citet{touvron2023llama2} allocate more resources to the prompt collection to maximize its diversity while featuring only two responses per prompt.
Conversely, some works introduce fine-grained distinctions to LLMs by incorporating list-wise comparisons among responses, or dynamically sampling better candidates for SFT~\cite{yuan2023rrhf, dong2023raft, song2023pro}, also leading to improved performance.

While more prompts can cover a wider range of domains and topics, limitations in annotation resources often force researchers to choose one side between diverse prompts and longer rankings with more responses. 
In our study, we investigate the impact of prompt diversity and compare it quantitatively with that of responses. We also establish empirical relations between prompt diversity and the final performance of tuned LLMs.

\subsection{Scaling Analyses of LLMs}
As LLMs continue to increase in scale, leading to higher training costs, it becomes crucial to make initial predictions regarding their performance.
Various key factors of LLMs can be scaled to predict the ultimate performance. 
From a micro perspective, \citet{kaplan2020scaling} and \citet{openai2023gpt} try to formulate power laws from the model size or the amount of computation of LLMs to their converged loss values during pre-training.
In contrast, \citet{lee2023rlaif} examine the impact of different training paradigms for human alignment from a macro perspective. 
Additionally, \citet{zhang2023wider} explore how the assembly of LLMs can influence the final performance, and \citet{yuan2023scaling} show that loss values can even indicate the accuracy of mathematical reasoning.

The impact of data used in the pre-training or fine-tuning stages can also be investigated. 
\citet{kaplan2020scaling} and \citet{muennighoff2023scaling} scale the total number of tokens associated with the performance levels achievable by LLMs
However, token count may not perfectly represent the diversity of data distribution. 
\citet{zhao2023preliminary} accordingly propose a tree-like structure for instruction alignment and study the scaling relationship between the complexity of instructions and final success rates. 
\citet{lu2023instag} and \citet{wei2023instructiongpt} propose different metrics for the estimation data diversity to label or filter training samples, building upon the observation~\citep{zhou2023lima} that a small dataset can unlock the specific capabilities of LLMs through fine-tuning. 
Building on these studies, we concentrate on the distribution of split prompts and responses for human alignment, and provide detailed analyses of performance improvement influenced by dataset sizes and diversity.
\section{Quantitative Experiments}
\subsection{Background}
Different from pre-training, individual samples are typically divided into a prompt $x$ and a response $y$ in LLM fine-tuning. 
Specifically in human alignment, a single prompt can be associated with multiple responses $y^{1:n} = y^1, y^2, \cdots, y^n$, ranked according to varying levels of preference, which are learned by LLMs to enhance their outputs.

Intuitively, a broad range of prompts can potentially enhance the generalization ability of LLMs, thus improving their final performance. 
Likewise, using diverse candidate responses can be beneficial by enabling LLMs to capture subtle distinctions reflecting different preferences.
It is difficult to determine the ideal number of samples and the optimal length of response rankings for LLMs to align with human preference. 
However, the fact is that human annotations are always costly, and the total amount of annotations can be accordingly limited.
Therefore, given a fixed amount of human annotations, there has to be a trade-off between increasing prompts~(while reducing the length of response rankings), or associating each prompt with more responses~(but fewer prompts in total) in the dataset. 
Researchers need to make a choice between these two directions~\citep{ouyang2022training, yuan2023rrhf, dong2023raft, song2023pro, touvron2023llama2}.

In this section, we design a quantitative experiment aimed at conducting preliminary comparisons of their effects.
We select a series of subsets for fine-tuning, all sharing the same total annotation volume, some emphasizing more prompts while others prioritize more responses.
Subsequently, we apply two well-known algorithms to these subsets and aggregate their performance results to assess the impact of different configurations.

\subsection{Dataset Construction}
Similar to ~\citet{yuan2023rrhf},~\citet{rafailov2023direct} and~\citet{song2023pro}, we utilize the \textbf{Human Preference Data on Helpfulness and Harmlessness}, referred to as \textit{HH-RLHF}~\citeplanguageresource{bai2022rlhfdata}, as the foundational dataset.
Each original sample consists of a common prompt and two candidate responses~(named 2-ranking), one chosen by human annotators and the other rejected.
We extend each 2-ranking into 4-ranking through zero-shot augmentation using Curie~\citep{brown2020language} and Alpaca~\citep{alpaca}, neither of which has been fine-tuned for human alignment previously.

Assuming the total volume of human annotations is 2N, there are various subsets with different prompt sizes and response ranking lengths. For example, each subset may consist of N prompts with 2-rankings, 2N/3 prompts with 3-rankings, or N/2 prompts with 4-rankings, which all maintain 2N annotations~(2N$=$2$\times$N$=$3$\times$2N/3$=$4$\times$N/2).

We also attach additional subsets containing N prompts, 2N/3 prompts, and N/2 prompts with 2/3/4-rankings, to present comprehensive results for further analyses.

\begin{table*}
\centering
\scalebox{0.96}{
  \begin{tabular}{llllllll}
    \toprule
    \multicolumn{1}{l}{\multirow{2}{*}{\begin{tabular}{l}\textbf{Settings} \\ (Algorithm, Backbone, Domain)\end{tabular}}} & \multicolumn{1}{c}{\multirow{2}{*}{\begin{tabular}{c}\textbf{\# Candidate} \\ \textbf{Responses}\end{tabular}}} & \multicolumn{3}{c}{\textbf{\# Prompts}~(N$=$24000)} & \multicolumn{3}{c}{\textbf{\# Prompts}~(N$=$3000)} \\ 
    \cmidrule(lr){3-5}\cmidrule(lr){6-8}
    \multicolumn{1}{l}{} & \multicolumn{1}{c}{} & \multicolumn{1}{c}{N/2} & \multicolumn{1}{c}{2N/3} & \multicolumn{1}{c}{N} & \multicolumn{1}{c}{N/2} & \multicolumn{1}{c}{2N/3} & \multicolumn{1}{c}{N} \\ \midrule
    \multicolumn{1}{l}{\multirow{3}{*}{\begin{tabular}{l}PRO, OPT-1.3B, Harmless\end{tabular}}} & \multicolumn{1}{c}{2} & \multicolumn{1}{c}{55.58} & \multicolumn{1}{c}{55.58} & \multicolumn{1}{c}{\textbf{\textcolor{red!70!black}{57.01}}} & \multicolumn{1}{c}{50.42} & \multicolumn{1}{c}{51.37} & \multicolumn{1}{c}{\textbf{\textcolor{red!70!black}{53.65}}} \\
    \multicolumn{1}{l}{} & \multicolumn{1}{c}{3} & \multicolumn{1}{c}{57.11} & \multicolumn{1}{c}{\textbf{\textcolor{red!70!black}{57.29}}} & \multicolumn{1}{c}{59.28} & \multicolumn{1}{c}{53.28} & \multicolumn{1}{c}{\textbf{\textcolor{red!70!black}{54.80}}} & \multicolumn{1}{c}{56.47} \\
    
    \multicolumn{1}{l}{} & \multicolumn{1}{c}{4} & \multicolumn{1}{c}{\textbf{\textcolor{red!70!black}{59.24}}} & \multicolumn{1}{c}{58.98} & \multicolumn{1}{c}{59.92} & \multicolumn{1}{c}{\textbf{\textcolor{red!70!black}{55.36}}} & \multicolumn{1}{c}{56.73} & \multicolumn{1}{c}{58.28} \\
    
    \midrule

    \multicolumn{1}{l}{\multirow{3}{*}{\begin{tabular}{l}PRO, OPT-1.3B, Helpful\end{tabular}}} & \multicolumn{1}{c}{2} & \multicolumn{1}{c}{49.05} & \multicolumn{1}{c}{49.09} & \multicolumn{1}{c}{\textbf{\textcolor{red!70!black}{50.06}}} & \multicolumn{1}{c}{44.69} & \multicolumn{1}{c}{45.37} & \multicolumn{1}{c}{\textbf{\textcolor{red!70!black}{46.49}}} \\
    
    \multicolumn{1}{l}{} & \multicolumn{1}{c}{3} & \multicolumn{1}{c}{49.98} & \multicolumn{1}{c}{\textbf{\textcolor{red!70!black}{51.00}}} & \multicolumn{1}{c}{51.43} & \multicolumn{1}{c}{45.74} & \multicolumn{1}{c}{\textbf{\textcolor{red!70!black}{47.01}}} & \multicolumn{1}{c}{49.73} \\
    
    \multicolumn{1}{l}{} & \multicolumn{1}{c}{4} & \multicolumn{1}{c}{\textbf{\textcolor{red!70!black}{51.35}}} & \multicolumn{1}{c}{51.04} & \multicolumn{1}{c}{51.74} & \multicolumn{1}{c}{\textbf{\textcolor{red!70!black}{48.85}}} & \multicolumn{1}{c}{48.69} & \multicolumn{1}{c}{50.33} \\
    
    \midrule

    \multicolumn{1}{l}{\multirow{3}{*}{\begin{tabular}{l}PRO, OPT-1.3B, Global\end{tabular}}} & \multicolumn{1}{c}{2} & \multicolumn{1}{c}{52.78} & \multicolumn{1}{c}{52.73} & \multicolumn{1}{c}{\textbf{\textcolor{red!70!black}{53.78}}} & \multicolumn{1}{c}{47.64} & \multicolumn{1}{c}{48.57} & \multicolumn{1}{c}{\textbf{\textcolor{red!70!black}{50.13}}} \\
    
    \multicolumn{1}{l}{} & \multicolumn{1}{c}{3} & \multicolumn{1}{c}{53.91} & \multicolumn{1}{c}{\textbf{\textcolor{red!70!black}{54.67}}} & \multicolumn{1}{c}{55.51} & \multicolumn{1}{c}{49.56} & \multicolumn{1}{c}{\textbf{\textcolor{red!70!black}{50.76}}} & \multicolumn{1}{c}{53.28} \\
    
    \multicolumn{1}{l}{} & \multicolumn{1}{c}{4} & \multicolumn{1}{c}{\textbf{\textcolor{red!70!black}{55.30}}} & \multicolumn{1}{c}{55.12} & \multicolumn{1}{c}{55.96} & \multicolumn{1}{c}{\textbf{\textcolor{red!70!black}{52.28}}} & \multicolumn{1}{c}{52.54} & \multicolumn{1}{c}{54.19} \\

    \midrule

    \multicolumn{1}{l}{\multirow{3}{*}{\begin{tabular}{l}SFT, OPT-1.3B, Global\end{tabular}}} & \multicolumn{1}{c}{2} & \multicolumn{1}{c}{52.25} & \multicolumn{1}{c}{52.78} & \multicolumn{1}{c}{\textbf{\textcolor{red!70!black}{52.63}}} & \multicolumn{1}{c}{49.85} & \multicolumn{1}{c}{49.33} & \multicolumn{1}{c}{\textbf{\textcolor{red!70!black}{50.47}}} \\
    
    \multicolumn{1}{l}{} & \multicolumn{1}{c}{3} & \multicolumn{1}{c}{53.60} & \multicolumn{1}{c}{\textbf{\textcolor{red!70!black}{54.18}}} & \multicolumn{1}{c}{54.20} & \multicolumn{1}{c}{51.59} & \multicolumn{1}{c}{\textbf{\textcolor{red!70!black}{51.26}}} & \multicolumn{1}{c}{51.35} \\
    
    \multicolumn{1}{l}{} & \multicolumn{1}{c}{4} & \multicolumn{1}{c}{\textbf{\textcolor{red!70!black}{55.06}}} & \multicolumn{1}{c}{55.00} & \multicolumn{1}{c}{56.27} & \multicolumn{1}{c}{\textbf{\textcolor{red!70!black}{51.93}}} & \multicolumn{1}{c}{51.97} & \multicolumn{1}{c}{53.03} \\

    \midrule

    \multicolumn{1}{l}{\multirow{3}{*}{\begin{tabular}{l}PRO, LLaMA-7B, Global\end{tabular}}} & \multicolumn{1}{c}{2} & \multicolumn{1}{c}{54.53} & \multicolumn{1}{c}{54.68} & \multicolumn{1}{c}{\textbf{\textcolor{red!70!black}{55.16}}} & \multicolumn{1}{c}{53.16} & \multicolumn{1}{c}{53.06} & \multicolumn{1}{c}{\textbf{\textcolor{red!70!black}{53.31}}} \\
    
    \multicolumn{1}{l}{} & \multicolumn{1}{c}{3} & \multicolumn{1}{c}{56.58} & \multicolumn{1}{c}{\textbf{\textcolor{red!70!black}{55.89}}} & \multicolumn{1}{c}{56.26} & \multicolumn{1}{c}{54.74} & \multicolumn{1}{c}{\textbf{\textcolor{red!70!black}{53.00}}} & \multicolumn{1}{c}{56.05} \\
    
    \multicolumn{1}{l}{} & \multicolumn{1}{c}{4} & \multicolumn{1}{c}{\textbf{\textcolor{red!70!black}{57.50}}} & \multicolumn{1}{c}{56.42} & \multicolumn{1}{c}{57.26} & \multicolumn{1}{c}{\textbf{\textcolor{red!70!black}{55.25}}} & \multicolumn{1}{c}{55.88} & \multicolumn{1}{c}{55.35} \\

    \midrule

    \multicolumn{1}{l}{\multirow{3}{*}{\begin{tabular}{l}SFT, LLaMA-7B, Global\end{tabular}}} & \multicolumn{1}{c}{2} & \multicolumn{1}{c}{53.18} & \multicolumn{1}{c}{53.29} & \multicolumn{1}{c}{\textbf{\textcolor{red!70!black}{54.63}}} & \multicolumn{1}{c}{52.78} & \multicolumn{1}{c}{53.03} & \multicolumn{1}{c}{\textbf{\textcolor{red!70!black}{53.80}}} \\
    
    \multicolumn{1}{l}{} & \multicolumn{1}{c}{3} & \multicolumn{1}{c}{56.01} & \multicolumn{1}{c}{\textbf{\textcolor{red!70!black}{55.53}}} & \multicolumn{1}{c}{56.07} & \multicolumn{1}{c}{53.98} & \multicolumn{1}{c}{\textbf{\textcolor{red!70!black}{53.74}}} & \multicolumn{1}{c}{54.50} \\
    
    \multicolumn{1}{l}{} & \multicolumn{1}{c}{4} & \multicolumn{1}{c}{\textbf{\textcolor{red!70!black}{56.13}}} & \multicolumn{1}{c}{56.30} & \multicolumn{1}{c}{57.30} & \multicolumn{1}{c}{\textbf{\textcolor{red!70!black}{55.05}}} & \multicolumn{1}{c}{55.00} & \multicolumn{1}{c}{56.14} \\
    
    \bottomrule
  \end{tabular}
}
\caption{\label{pilot_pro_res}
    Results of quantitative experiments. LLMs can acquire better performance with either diverse prompts or responses used for fine-tuning, while increasing responses benefits LLMs more than increasing prompts with the same amount of annotations~(highlighted in red bold).
}
\end{table*}

\subsection{Metrics}
Unlike some tasks that can be easily measured, human preference can be more abstract and hard to estimate. 
Both \citet{yuan2023rrhf} and \citet{song2023pro} utilize RMs to evaluate the performance of fine-tuned LLMs, while the emerging GPT-4-as-a-judge in human alignment~\citep{rafailov2023direct} can be also convincing.
Our evaluation predominantly relies on public reward models, employing distinct reward models RM$_{\text{train}}$ and RM$_{\text{test}}$ for training and testing phases, respectively. 
The outcomes are then cross-validated by GPT-4 assessments. 
For fine-tuning, we combine all 4 subsets in \textit{HH-RLHF} for LLM fine-tuning and present the outcomes for two representative subsets, namely \textit{Harmless}$_{\text{base}}$ and \textit{Helpful}$_{\text{base}}$. Furthermore, we provide the overall scores for all test samples across the 4 subsets.

\subsection{Benchmark Algorithms}
For each dataset, we select representative supervised methods as benchmark algorithms, because supervised training can directly reflect the impact of the datasets involved. 
Specifically, we opt for two widely-used algorithms, namely \textbf{Supervised Fine-tuning} and \textbf{Preference Ranking Optimization}~\cite{song2023pro}, denoted SFT and PRO, to represent other methods that are either sensitive or insensitive to response rankings.

To elaborate, SFT is similar to the pre-training process but exerts supervision solely on the top candidate $y^1$,
\begin{equation}
\label{eq:sft}
    \mathcal{L}_{\text{SFT}}(y^1 \mid x) = -\sum_{i=1}^{\left | y^1 \right |} \log_{}{p_{\Theta}\left(y^1_i\middle|x, y^1_{<i}\right)} 
\end{equation}
Instead, PRO forces the LLM to distinguish the best one from multiple candidates. It utilizes the whole ranking $y^1, y^2,$ $\cdots, y^n$ through multiple one-to-many contrasts, implemented as:\\
\begin{equation}
    \mathcal{L}(y^{1:n} \mid x) = -\sum_{k=1}^{n-1} \log \frac{\exp\left( \frac{\log p_{\Theta}(x, y^k)}{\mathcal{T}^k_k} \right)}{\sum\limits_{i=k}^{n} \exp\left( \frac{\log p_{\Theta}(x, y^i)}{\mathcal{T}^i_k} \right)}
\end{equation}
\begin{equation}
    \mathcal{T}^{i>k}_k = \frac{1}{r_{\phi}(x, y^k) - r_{\phi}(x, y^i)}
\end{equation}
\begin{equation}
    \mathcal{T}^k_k = \min_{i>k} \mathcal{T}^{i}_k
\end{equation}
and the final objective appends $\mathcal{L}_{\text{SFT}}$ for a balance between text quality and human preference,
\begin{equation}
\label{eq:pro}
    \mathcal{L}_{\text{PRO}}(y^{1:n} \mid x) = \beta\mathcal{L}_{\text{SFT}}(y^1 \mid x) + \mathcal{L}(y^{1:n} \mid x) 
\end{equation}

\subsection{Implementation Details}
The experiments are conducted with different $N$~(24000 and 3000).
We mainly utilized OPT-1.3B~\cite{zhang2022opt} as the base LLM and tested it with three different seeds, while incorporating LLaMA-7B~\cite{touvron2023llama} with just one seed due to computational constraints.
For the fine-tuning process of LLMs, we configured the total training steps to 4000 for each dataset, performing validation every 500 steps. 
Both RM$_{\text{train}}$\footnote{\url{https://huggingface.co/OpenAssistant/oasst-rm-2.1-pythia-1.4b-epoch-2.5}} and RM$_{\text{test}}$\footnote{\url{https://huggingface.co/OpenAssistant/oasst-rm-2-pythia-6.9b-epoch-1}} are publicly available checkpoints. 

The original dataset comprises newly added data without human annotations. Therefore, we first score all responses using RM$_{\text{train}}$, then re-rank them based on their scores. 
Furthermore, we ensure that datasets of larger sizes will encompass their smaller counterparts with rankings of the same length, while for datasets of the same size, longer rankings will include the shorter ones for each sample. 
More details are available in the code\footnote{\url{https:github.com/F2-song/ScalingAlignment}}. 

\begin{figure}[t]
\centering
\includegraphics[width=\linewidth]{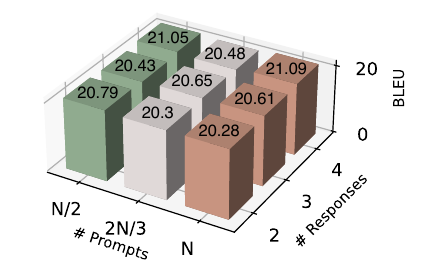}
\caption{Distribution of BLEU scores with different settings.}
\label{fig:bleu}
\end{figure}
\subsection{Results of Automatic Evaluation}
As demonstrated in Table~\ref{pilot_pro_res}, we gathered the average reward scores corresponding to various settings, including algorithms, backbones, and domains. These results aim to address the following research questions~(RQs).

\subsubsection*{RQ1: More Diverse Prompts or Responses?}
Longer response rankings and more prompts are both beneficial, but their effects are different. 
Generally, with an equal quantity of annotations, extending each ranking of responses leads to \textbf{better} enhancement compared to expanding prompts, regardless of the backbones used or the values of N.
These outcomes are highlighted in red bold within each grid, where scores in the \textbf{lower left} side indicate longer response rankings but fewer prompts, while scores in the \textbf{upper right} side represent the opposite scenario.
This observation is compatible with the hypothesis that LLMs possess the potential for human alignment that can be activated with a small number of samples~\citep{zhou2023lima}.
However, more responses for each prompt offer clearer alignment signals through comparisons, leading to a more significant optimization of LLMs.

In detail, an increased number of responses benefits PRO more than SFT, as the former emphasizes the importance of response rankings lacking in the latter.
Nonetheless, a longer ranking represents more samplings from the whole linguistic space, where the preferred response is more likely to be identified.
This explains why SFT methods also benefit from expanding responses.
Further examination of the results in the domains of harmlessness and helpfulness with OPT-1.3B and PRO confirms the validity of the above statement.

\subsubsection*{RQ2: How Does the Allocation of Annotations Impact the Quality of Output Texts?}
Changes in the allocation of annotations~(more prompts or more responses) appear to have no impact on the outcome.
We hereby plot the BLEU distribution generated by the OPT-1.3B model fine-tuned with PRO and 24000 samples in Figure~\ref{fig:bleu}, where the distribution does not present a consistent pattern but fluctuates randomly across different configurations. 
The variations observed may be explained by the fact that LLMs possess robust language modeling abilities, consequently not requiring too many samples.
It underscores the importance of allocating more annotation resources to prepare responses when a certain number of prompts have been guaranteed.

\subsubsection*{RQ3: How Many Samples Are Sufficient for Human Alignment Fine-tuning?}
Intuitively, the more samples for fine-tuning, the more diverse they are, leading to potentially greater improvements for the tuned LLMs. 
However, determining the adequate quantity is a complex task, as it depends on factors like the algorithms, base models, and the number of responses.
For instance, while LLaMA-7B demonstrates notably high scores with 3000 samples, surpassing OPT-1.3B with an equal amount of training data, it shows a slower increase in performance compared to OPT-1.3B when more samples are included. 
Moreover, the degree of improvement achieved by increasing the dataset size usually decreases.
This is because the dataset is more likely to contain duplicate or similar content as the number of samples grows, making it less efficient to continually invest resources in comparison to potential performance gains.
\begin{figure*}[t]
\centering
\includegraphics[width=\textwidth]{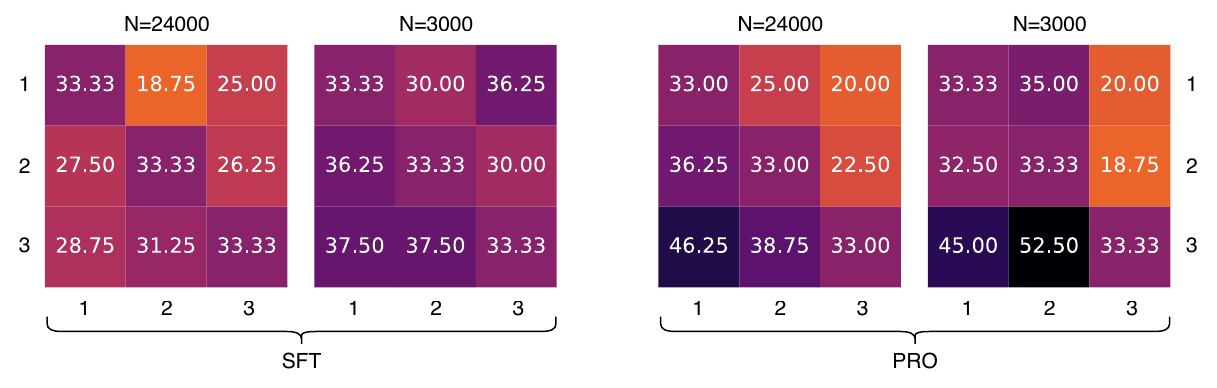}
\caption{GPT-4 Evaluation}
\label{fig:gpt4}
\end{figure*}
\subsection{GPT-4 Evaluation}
Apart from the automatic evaluation above with RM$_\text{test}$, we also take GPT-4-as-a-judge into consideration, since it has been widely recognized as an efficiently human-like tool to give fair judgment, especially for abstract concepts like human preference~\citep{zheng2023judging, song2023pro, dubois2023alpacafarm}. 
It further validates the statistical findings in Table~\ref{pilot_pro_res} by directly comparing the three settings as described below:\\
\textbf{Setting 1:} N prompts, i.e., a total of N samples, each with 2 responses.\\
\textbf{Setting 2:} 2N/3 prompts, each with 3 responses.\\
\textbf{Setting 3:} N/2 prompts, each with 4 responses.\\
All three settings have a total of 2N annotations for fine-tuning. 

We use LLaMA-7B for fine-tuning under these three settings since it can yield high-quality outputs. 
We randomly select prompts from the test sets of \textit{Harmless}$_{\textit{base}}$ and \textit{Helpful}$_{\textit{base}}$ for evaluation. 
The outputs of each tuned LLaMA are compared with those under other settings, scored directly by GPT-4 through bi-directional comparisons to enhance fairness~\cite{wang2023eval}, and the win-lose rates of each comparison are depicted in Figure~\ref{fig:gpt4}. 
In each matrix $M$, each row $i$~(or column $j$) corresponds to Setting $i$~(or Setting $j$), with the element $M_{i,j}$ indicating the win rate of LLM outputs tuned with Setting $i$ against those with Setting $j$.
The diagonal elements in $M$ are uniformly set at 33.33 for comparisons between two identical contents, distributed evenly among [Win, Lose, Tie].

Figure~\ref{fig:gpt4} illustrates that $M_{i,j}$ always surpasses $M_{j,i}$ along the main diagonal. This implies that the win rate of Setting $i$ against Setting $j$ is always higher than its loss rate, consistent with the results in Table~\ref{pilot_pro_res}. 
This also proves that RM$_{\text{test}}$ can be a reliable evaluator. 
In general, it reaffirms the conclusion that increasing annotations for responses improves LLMs to better align with human preference than prompts.
\section{The Scaling Law between Prompt Diversity and LLMs Preference}
While the diversity of prompts or responses can both be beneficial to LLMs fine-tuning, increasing prompts is less effective compared to increasing responses.
This difference can be attributed to the inefficiency of using quantity alone to measure prompt diversity.
In this section, we explore the concept of prompt diversity.
We first discuss the importance of controlling prompt diversity and then propose a new empirical formulation for it.
Additionally, assuming all other factors remain constant~(such as base models, fine-tuning algorithms, annotation sources, and lengths of response rankings), a \textbf{linear correlation} between the final performances of fine-tuned LLMs and the calculated diversity from various subsets can be illustrated.

\begin{figure*}[t]
\centering
\includegraphics[width=\textwidth]{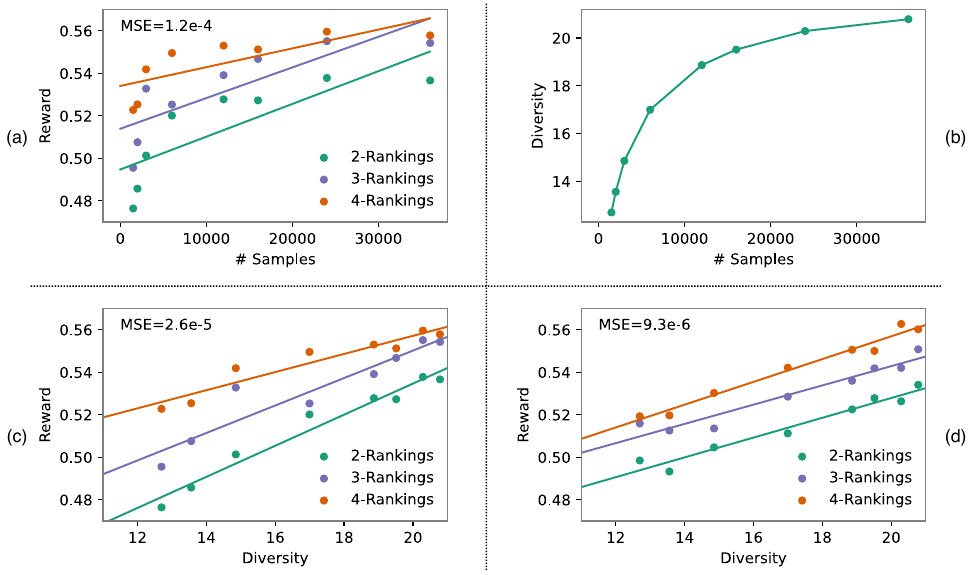}
\caption{(a)~Linear fitting from different sample amounts to finally acquired rewards of LLMs tuned with \textbf{PRO}. (b)~The trend of diversity with the increasing sample amount. (c)~Linear fitting from the proposed diversity metric to finally acquired rewards of LLMs tuned with \textbf{PRO}. (d)~Linear fitting from the proposed diversity metric to finally acquired rewards of LLMs tuned with \textbf{SFT}.}
\label{fig:diversity}
\end{figure*}
\subsection{Diversity Formulation}
Utilizing quantity to control response diversity appears to be rational. 
Given that new responses typically originate from various sources, and the range of response rankings is relatively limited, augmenting the number of responses can lead to significant variations. 
The quantitative experiments also demonstrate that expanding response rankings contributes to improvements for LLMs.

However, the approach becomes oversimplified when applied to prompt diversity. 
Here the diversity is based on all prompts, and minor adjustments in quantity may not have a noticeable impact, as evidenced by the marginal improvement from N/2 to 2N/3~(for N=3000) in the quantitative experiments, while potential duplication in prompts can be another factor. 
Furthermore, fine-grained features within utterances, such as semantics, contexts, and even syntax, are crucial to prevalent LLMs that depend on tokenization followed by causal modeling. 
They should also be taken into consideration.

Moreover, the redundant prompts mentioned earlier do not contribute significantly to enhancing overall diversity and should be initially removed. 
This can be measured by assessing the proportion of distinct N-grams within the dataset.

Different from \citet{kaplan2020scaling} and \citet{muennighoff2023scaling}, we leverage N-grams instead of individual tokens as the basic element for calculation, because N-grams inherently capture contextual details beyond the meaning of single tokens. 
In addition, the aforementioned duplicated data essentially do not provide any extra advantages in terms of overall diversity and should be filtered out initially. 
We define it as the rate of unique N-grams present in the dataset,
\begin{equation}
\label{eq:unique}
    r_{\text{unique}} = \frac{\left|\text{Filter}(G) \right|}{\left|G \right|}
\end{equation}
where $G$ is the collection of all N-grams derived from the tokenized corpus and $\text{Filter}(G)$ denotes the removal of repeated elements. 
Subsequently, the diversity metric $d$ can be formulated as product of $r_{\text{unique}}$ and the total number of prompts $m$,
\begin{equation}
    d = r_{\text{unique}} * m
\end{equation}

Empirically, as the number of prompts increases, the marginal effect decreases gradually, and the diversity should follow the same pattern. 
Therefore, we introduced a decay index $p$ to the sample quantity $m$ to incorporate decay into its growth rate. The concept of prompt diversity is formulated as,
\begin{equation}
\label{eq:formulation}
    d = r_{\text{unique}} * m^{p}
\end{equation}

\subsection{Analysis}
To examine the connection between the pre-defined diversity metric and the final performance of fine-tuned LLMs, we use 2-grams for calculation and collect \{1500, 2000, 3000, 6000, 12000, 16000, 24000, 36000\} samples from the original dataset with rankings of lengths 2/3/4. 
This resulted in 24 subsets for LLMs fine-tuning.
In this part, we empirically set the value of $p$ to 0.5 for \textit{HH-RLHF}, although it could also be found using grid search.
\begin{algorithm}
\caption{Determining the decay index with grid search.}
\label{algorithm:grid}
\LinesNumbered
\KwIn{
    Fine-tuning algorithm \textit{FT},
    Datasets~$\{D_i\}$ of ascending sizes,
    Language Model~$\pi$, 
    Step Length~$l$
}
\KwOut{
    Decay Index~$p$
}

\tcp{Fine-tuning and evaluation}
Let $S$ be an empty set

\For {$D_i \in \left \{D_1, ..., D_{n}\right \}$}{
     Let $\pi_i = \textit{FT}(\pi,D_i)$
     
     Evaluate $\pi_i$

     Let $s_i$ be the performance of $\pi_i$

     Add $s_i$ to $S$
}

\tcp{Searching the decay index}
Let $p = 0$, $\hat{p} = 0$, $L = \textit{inf}$

\While {$\hat{p}<1$}
{
    Let $\hat{p} = \hat{p} + l$
    
    Compute diversity degrees $\{d_i\}$ based on $\hat{p}$, $\{D_i\}$ and Equation~\ref{eq:formulation}

    Compute the MSE $\hat{L}$ using linear fitting on $\{d_i\}$ and $S$

    \If{$\{d_i\}$ is ascending \&\& $\hat{L}<L$}{
        Let $L=\hat{L}$, $p=\hat{p}$
    }
}
\textbf{return} $p$

\end{algorithm}

We start our analysis by visualizing the results of above 24 subsets in Figure~\ref{fig:diversity}(a). 
A discernible positive correlation between enhanced performance and the increasing quantity is observed. 
Furthermore, improved scores can also be achieved with longer response sequences.
Nevertheless, the growth in reward scores and quantity of prompts is not synchronized.
The former shows a gradual decline in speed, while the latter maintains a consistent pace. 
More precisely, the performance experiences a sharp increase with a rising number of samples at the beginning, yet tends to plateau with a larger volume of samples. 
Even when we convert the X-axis from sample quantity to token quantity, this conclusion still holds. 

A commonly accepted concept is that the diversity of a dataset may not continue to increase indefinitely. 
As the size of the dataset expands, new content often contains complete or partial duplications of earlier material. 
By analyzing actual datasets, we have graphed in Figure~\ref{fig:diversity}(b) the evolution of diversity outlined in Equation~\ref{eq:formulation} as the sample size grows. 
This graph aligns with the idea that the rate of diversity growth should gradually decrease. 
Based on the similar patterns observed in performance and diversity trends, a linear correlation between these them may exist,
\begin{equation}
    r = \alpha * d + \beta + \epsilon
\end{equation}
where $\alpha$ and $\beta$ are coefficients, while $r$ and $\epsilon$ denote the reward score and error term, respectively. 

We then gather the performance~(i.e., average reward) achieved by OPT-1.3B under the supervision of PRO and SFT, respectively, and then apply linear fitting to correlate each final score with the computed diversity of the specific subset used. 
The outcomes are presented in Figure~\ref{fig:diversity}(c) and (d). 
Additionally, we compare these results with the linear fitting between performance and the sample quantity, as shown in Figure~\ref{fig:diversity}(a), while it leads to a significantly higher mean squared error~(MSE) of 1.2e-4 compared to 2.8e-5 and 8.8e-6 corresponding to Figure~\ref{fig:diversity}(c) and (d), respectively. 
This validates the linear correlation between our proposed $d$ and the final performance.
We also compute the MSE values with LLaMA-7B, which are marginally higher than those with OPT-1.3B, possibly due to fluctuations in a single seed~(4.6e-5 for PRO and 1.7e-5 for SFT).
\section{Sampling with Diversity Check}
In this section, we present a technique for data augmentation using the existing samples.
Fresh samples are first collected and then selected to enrich the overall variety of prompts. This selection is aimed to optimize the local diversity between the new and existing samples. 
We initially demonstrate the effectiveness of this technique in enhancing prompt diversity. 
By fine-tuning LLMs on the augmented datasets, there is also a slight performance enhancement along with increased diversity.

\subsection{Augmentation}
We design the data augmentation as where there are existing samples that constitute a seed set.
In this setting, $n$ samples are randomly selected to support each augmentation iteration.
To simplify the experiment, we reuse one subset $D$ in the last section as the aforementioned seed set, and select new samples from the original \textit{HH-RLHF} to simulate the process of augmentation.
It is ensured that each new sample is selected from the non-overlapping portion of \textit{HH-RLHF} concerning $D$.

\subsection{Filtering with Diversity}
We start by revisiting the concept of prompt diversity. 
The proposed metric above can be affected by two factors: $r_{\text{unique}}$, representing the ratio of unique N-grams, and the total number of prompts, which rises with decreased speed. 
Consequently, by simultaneously increasing $r_{\text{unique}}$ during data augmentation, the diversity metric can experience a more rapid growth.

However, identifying a batch of new samples that maximizes the diversity of the total $D$ can be challenging. 
Therefore, we introduce a locally greedy search process to filter new samples based on the supporting samples. 
Specifically, by computing the Jaccard Index between the set $X$ of supporting samples and the $i$-th element $Y_i$ of set $Y$,
\begin{equation}
    \text{Jaccard}_i = \frac{\text{Filter}(G_X) \cap \text{Filter}(G_{Y_i})}{\text{Filter}(G_X) \cup \text{Filter}(G_{Y_i})}
\end{equation}
where $Y_k$ with the lowest $\text{Jaccard}_i$ can be selected to enhance the local $r_{\text{unique}}$, thereby boosting the overall diversity.

\begin{table}
\centering
\scalebox{0.94}{
  \begin{tabular}{lllll}
    \toprule
    \multicolumn{1}{l}{\textbf{Method}} & \multicolumn{1}{l}{\textbf{\# Responses}} & \multicolumn{1}{c}{$D_{12000}$} & \multicolumn{1}{c}{\Large{\arrowmark}} & \multicolumn{1}{c}{$\hat{D}_{12000}$} \\
    \midrule
    \multicolumn{1}{l}{\multirow{3}{*}{PRO}} & \multicolumn{1}{c}{2} & \multicolumn{1}{c}{52.78} & \multicolumn{1}{c}{} & \multicolumn{1}{c}{52.85} \\
    \multicolumn{1}{l}{} & \multicolumn{1}{c}{3} & \multicolumn{1}{c}{53.91} & \multicolumn{1}{c}{\Large{\arrowmark}} & \multicolumn{1}{c}{54.55} \\
    \multicolumn{1}{l}{} & \multicolumn{1}{c}{4} & \multicolumn{1}{c}{55.30} & \multicolumn{1}{c}{} & \multicolumn{1}{c}{55.50} \\
    \midrule
    \multicolumn{1}{l}{\multirow{3}{*}{SFT}} & \multicolumn{1}{c}{2} & \multicolumn{1}{c}{52.25} & \multicolumn{1}{c}{} & \multicolumn{1}{c}{51.81} \\
    \multicolumn{1}{l}{} & \multicolumn{1}{c}{3} & \multicolumn{1}{c}{53.60} & \multicolumn{1}{c}{\Large{\arrowmark}} & \multicolumn{1}{c}{54.47} \\
    \multicolumn{1}{l}{} & \multicolumn{1}{c}{4} & \multicolumn{1}{c}{55.06} & \multicolumn{1}{c}{} & \multicolumn{1}{c}{55.20} \\
    \midrule
    \multicolumn{2}{l}{\textbf{Diversity $d$ of Prompts}} & \multicolumn{1}{c}{18.86} & \multicolumn{1}{c}{\Large{\arrowmark}} & \multicolumn{1}{c}{19.75} \\
    \bottomrule
  \end{tabular}
}
\caption{\label{aug_res}
    Results of Data Augmentation.
}
\end{table}
\subsection{Results}
We set $n$ as 2 and utilize the subset $D_{6000}$ containing 6000 samples~(for all versions with 2/3/4 rankings), to which 6000 new samples are then added, forming $\hat{D}_{12000}$.
Meanwhile, we treat the subset $D_{12000}$ as its counterpart, comprising 12000 randomly sampled samples without filtration and covering $D_{6000}$. 
We apply PRO and SFT on OPT-1.3B using these subsets and present the outcomes in Table~\ref{aug_res}. 
The diversity of prompts increases from 18.86 to 19.75, and slight enhancements are shown with few exceptions, which can be normal fluctuations. 
This primarily demonstrates the impact of the proposed filtering method. 
It may potentially be amplified with larger $n$, a direction we leave for future research.
\section{Conclusion}
\label{sec:conclusion}
This study focuses on the impact of data diversity on human alignment fine-tuning.
Given the common limitation of available annotations in most scenarios, we investigate the effect of distributing them to enhance diversity in different ways, such as increasing prompts or responses. 
Our extensive experiments show that increasing the number of responses generally leads to greater enhancements in human alignment compared to expanding prompts. 
Additionally, we design an empirical metric to measure prompt diversity and reveal a linear correlation between it and the final performance of LLMs.
Finally, we propose a straightforward method to boost diversity in data augmentation, resulting in better performance of fine-tuned LLMs.
\section{Ethical Statement}
The presence of sensitive and offensive content within the datasets should be acknowledged. 
It is important to highlight that these contents do not reflect our views or beliefs, but are solely intended for research purposes.

Moreover, Curie inference and GPT-4 evaluation are utilized where they are available to adhere to legal requirements.
\section{Acknowledgement}
This work was supported by National Science and Technology Major Project (2022ZD0116308) and National Natural Science Foundation of China (62036001).

\nocite{*}
\section{Bibliographical References}\label{sec:reference}
\bibliographystyle{lrec-coling2024-natbib}
\bibliography{custom}

\section{Language Resource References}
\label{lr:ref}
\bibliographystylelanguageresource{lrec-coling2024-natbib}
\bibliographylanguageresource{languageresource}

\end{document}